\documentclass[10pt,twocolumn,letterpaper]{article}

\usepackage[pagenumbers]{iccv} 

\usepackage{algorithm}
\usepackage[noend]{algpseudocode}
\usepackage{multirow}
\usepackage{booktabs}
\usepackage{xcolor}
\usepackage{amssymb}
\usepackage{amsmath}
\usepackage{graphicx}
\usepackage{newfloat}
\usepackage{listings}
\definecolor{iccvblue}{rgb}{0.21,0.49,0.74}
\usepackage[pagebackref,breaklinks,colorlinks,allcolors=iccvblue]{hyperref}

\def\paperID{9860} 
\def\confName{ICCV}
\def\confYear{2025}

\title{The Key of Parameter Skew in Federated Learning}

\author{
Junfeng Liao\thanks{These authors contributed equally.},
Sifan Wang\footnotemark[1],
Ye Yuan,
Riquan Zhang\thanks{Corresponding author.} \\
Shanghai University of International Business and Economics\\
{\tt\small \{23349089, 23349111, yuany, rqzhang\}@suibe.edu.cn}
}

\begin{document}
\maketitle
\begin{abstract}
Federated Learning (FL) has emerged as an excellent solution for performing deep learning on different data owners without exchanging raw data. However, statistical heterogeneity in FL presents a key challenge, leading to skewness in local model parameter distributions that researchers have largely overlooked. In this work, we propose the concept of \textbf{parameter skew} to describe the phenomenon that can substantially affect the accuracy of global model parameter estimation. Additionally, we introduce \textbf{Fed}erated \textbf{Pa}rameter S\textbf{ke}w Learning (FedPake), a novel aggregation strategy to obtain a high-quality global model to address the implication from  \textit{parameter skew}. Specifically, we categorize parameters into high-dispersion and low-dispersion groups based on the coefficient of variation. For high-dispersion parameters, Micro-Class and Macro-Class represent the dispersion at the micro and macro levels, respectively, forming the foundation of FedPake. To evaluate the effectiveness of FedPake, we conduct extensive experiments with different FL algorithms on three Computer Vision datasets. FedPake outperforms eight state-of-the-art baselines by about 4.7\% in test accuracy.
\end{abstract}    
\section{Introduction}
\label{sec:intro}
\begin{figure}
    \centering
    \includegraphics[width=1\linewidth]{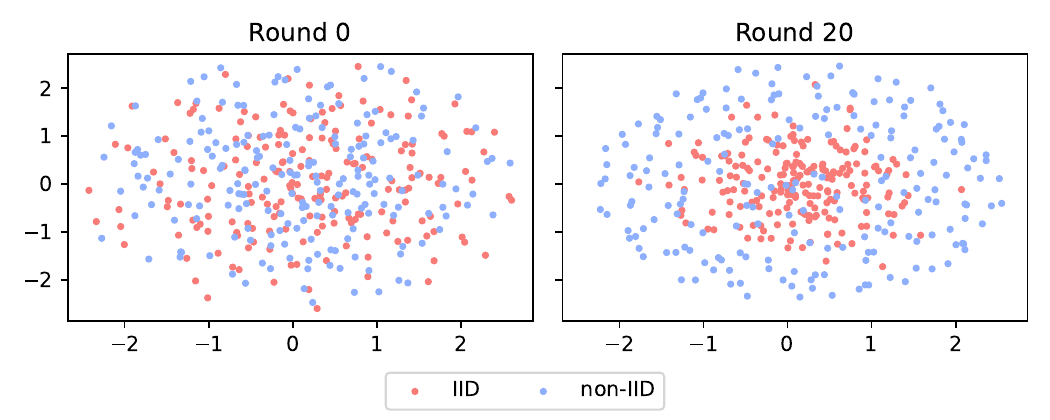}
    \caption{ \textbf{T-SNE visualizations illustrate the changes in the distribution of local model parameters during training under both IID and non-IID.} Under IID, parameters gradually converge during training, whereas under non-IID, they remain scattered. The experiments are conducted with ResNet-18 \cite{he2016deep} on CIFAR-10 dataset. }
    \label{fig:motivation}
\end{figure}

Federated Learning (FL) is a classical paradigm of distributed training that mitigates the communication barriers between datasets of different clients while enabling synchronous training with ensured data privacy \cite{shokri2015privacy,yang2019federated}. With the increasing attention to data privacy issues in the industry, FL has been widely applied in fields such as medicine and the Internet \cite{yang2021medicine,wu2020fedhome,liu2020fedvision}. FL has become an important and widely researched area in Machine Learning. FedAVG \cite{mcmahan2017communication}, a fundamental algorithm in FL, aggregates trained local models that are transmitted to the server in each round to update the global model, while the raw data from clients is not exchanged. A key challenge in FL is the heterogeneity of data distribution among different parties \cite{chen2022towards,shang2022federated,t2020personalized}. In the real world, data among parties can be non-Independent and Identically Distributed (non-IID), which makes dispersion among parameters of clients' models be enlarged during training, as shown in Figure \ref{fig:motivation}. Due to the dispersion, the global model may deviate from the optimal solution after aggregating local models from clients\cite{karimireddy2020scaffold}.

Several traditional federated learning (tFL) and personalized federated learning (pFL) studies have been conducted to address the non-IID issue during the training phase of local models \cite{li2020fedPROX,wang2020favor,wang2020fedma,kairouz2021advances,li2021ditto,oh2021fedbabu,zhang2023fedala}. For instance, FedProx \cite{li2020fedPROX} constrains the updates of local models using the $\ell_{2}\mathrm{-norm~}$ distance, while FAVOR \cite{wang2020favor} selects a subset of clients in each training round. Furthermore, FedMA \cite{wang2020fedma} utilizes statistical methods to alleviate data heterogeneity. Ditto \cite{li2021ditto} derives personalized models by incorporating regularization terms for each client to leverage information from the global model. FedBABU \cite{oh2021fedbabu} fine-tunes the classifier of the global model to obtain personalized models for individual clients. FedALA \cite{zhang2023fedala} proposes an adaptive aggregation strategy, enabling personalized models to selectively absorb information from the global model. However, these methods are unable to concentrate on the discrepancies among parameters of local models while focusing on the architecture of the model instead.

\begin{figure*}[ht]
    \centering
    \includegraphics[width=1\linewidth]{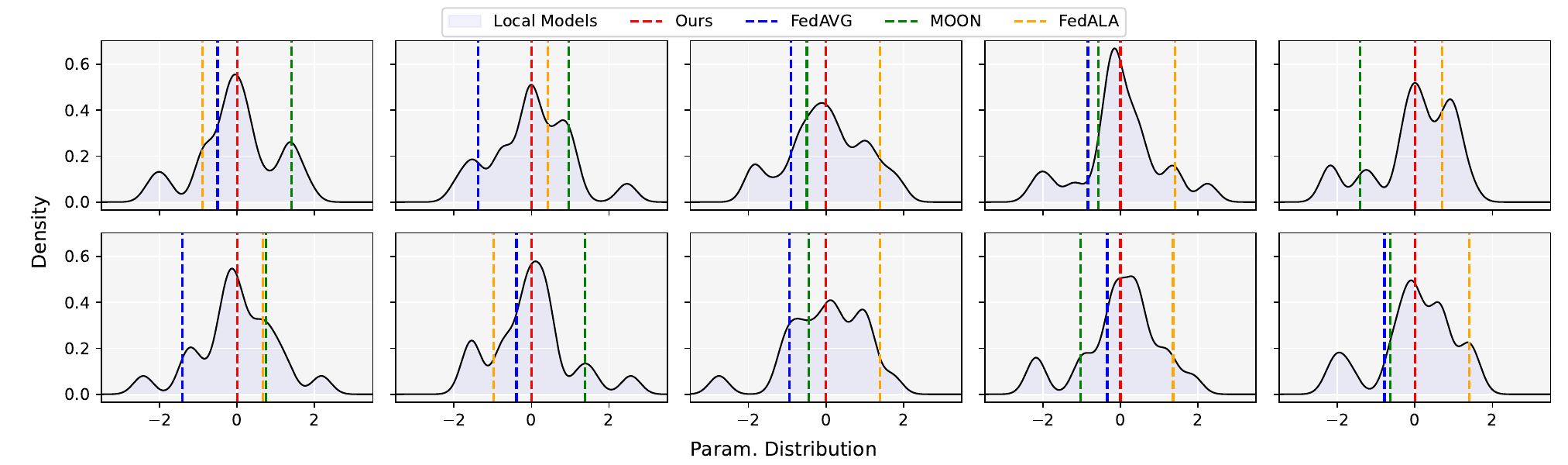}
    \caption{\textbf{The distribution of parameters of local models, {\color{red} FedPake(ours)}, {\color{green}{MOON}}, {\color{orange}{FedALA}}, and {\color{blue} FedAVG}.} The parameters in the figure are from ResNet-18 trained with FedPake and other methods on CIFAR-10. In the figure, the local model parameter's distribution is skewed, which clearly illustrates the presence of \textit{parameter skew}. Our method aligns closely with the main peak of the distribution, indicating that FedPake effectively captures the central tendency under \textit{parameter skew}. In contrast, other methods fail to address this issue, resulting in a deviation from the main peak.} 
    \label{fig:analysis}
\end{figure*}

\begin{figure*}[ht]
    \centering
    \includegraphics[width=1\linewidth]{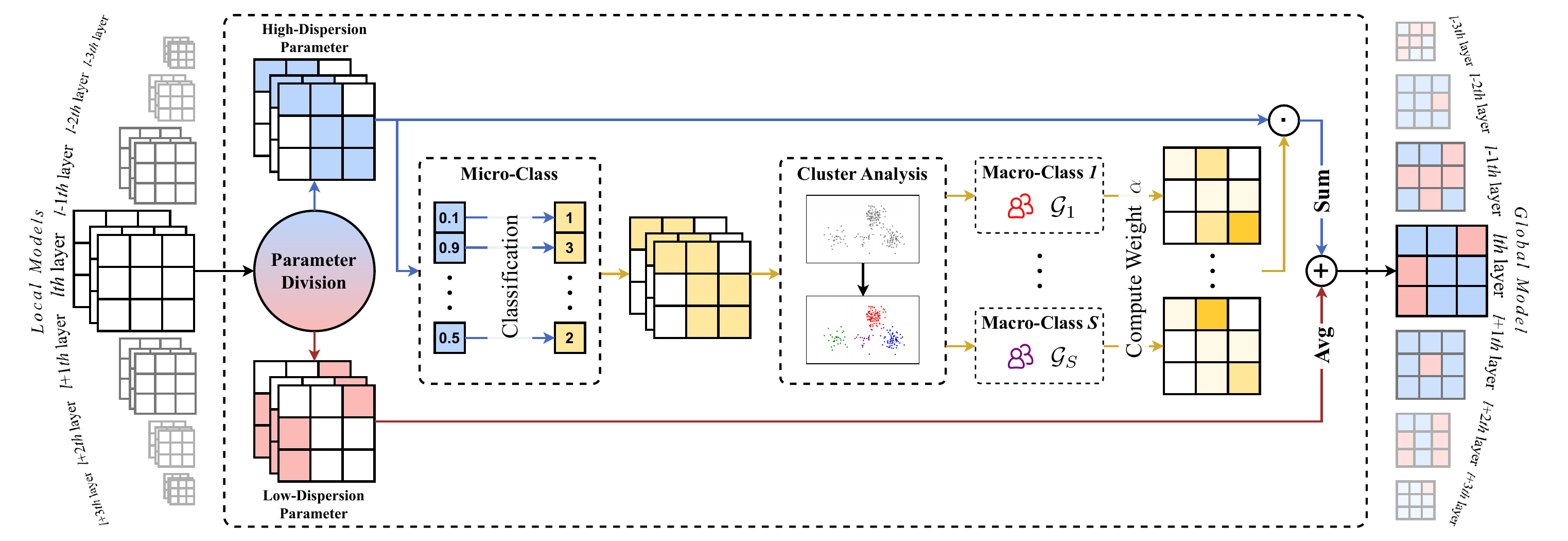}
    \caption{\textbf{The architecture of FedPake.} We input the local models into Parameter Division to obtain high-dispersion and low-dispersion parameters. For the high-dispersion, we calculate the final values using a weighted average, while average values serve as final values for the low-dispersion. Specifically, for the high-dispersion, our method computes weight $\alpha$ of each based on Micro-Class and Marco-Class. 3x3 convolutional kernel backbone are an instance.}
    \label{fig:architecture}
\end{figure*}

Our method is based on an observation, namely \textit{parameter skew} we proposed: as shown in Figure \ref{fig:motivation}, t-SNE \cite{van2008visualizing} is used to visualize the distribution of local model parameters, revealing that the non-IID issue causes dispersion among parameters of models with the same structure trained on different clients. Owing to the varying label distributions of data across different clients, models with identical structures learn disparate information on different clients, which leads to considerable variations in certain parameter values among local models. However, according to the Law of Large Numbers \cite{hsu1947complete}, skewness in the sample distribution can introduce significant bias into estimators, such as the mean, thereby reducing its robustness. The process of deriving a global model from local models can be seen as a parameter estimation task, where \textit{parameter skew} can significantly affect global model parameters. As illustrated in Figure \ref{fig:analysis}, the distribution of local models' parameters shows obvious skewness. Furthermore, FedAVG averages local model parameters to estimate the global model, potentially causing them to deviate from the central tendency and thereby weakening the model's robustness. To tackle the problem, we propose a novel FL algorithm, FedPake, shown in Figure \ref{fig:architecture}. We leverage the coefficient of variation \cite{brown1998coefficient} to categorize parameters into high-dispersion and low-dispersion. We continue to use the FedAVG process for parameters with low dispersion, while for high-dispersion parameters, we measure the extent of dispersion from both micro and macro perspectives, resulting in Micro-Class and Macro-Class. We assign weights to parameters according to the Micro-Class and Macro-Class, accounting for varying degrees of dispersion, to construct the global model.

To evaluate the effectiveness of FedPake, we conducted a comparative analysis with eight FL algorithms on CIFAR-10/100 \cite{krizhevsky2009learning} and Tiny-ImageNet \cite{chrabaszcz2017downsampled} datasets. The results, presented in Table \ref{tab:comparison}, indicate that our method surpasses other state-of-the-art (SOTA) methods. Additionally, in Figure \ref{fig:analysis}, we present the distribution of each parameter to illustrate the effectiveness of FedPake. In summary, our contributions are as follows:
\begin{itemize}
    \item We propose the \textit{parameter skew} resulting from heterogeneity and analyze its implications for the global model in FL.
    \item We introduce a novel FL algorithm, FedPake, which addresses the non-IID issue by leveraging \textit{parameter skew} to obtain Micro-Class and Macro-Class. Additionally, we analyze the effectiveness of FedPake and elucidate the reasons why other FL methods fail to achieve optimal performance.
    \item We conducted comprehensive experiments comparing FedPake with other baseline methods using three widely used datasets. FedPake outperformed eight SOTA methods, achieving up to a 4.7\% improvement in test accuracy while incurring lower computational costs.
\end{itemize}

\section{Related Work}
\subsection{Traditional Federated Learning} 
The traditional federated learning model, FedAVG \cite{mcmahan2017communication}, derives a global model through the aggregation of local models. Nevertheless, the heterogeneity of data across different clients detrimentally affects the performance of FedAVG. To address this challenge, FedProx \cite{li2020fedPROX} enhances the stability and generalization capability of the algorithm by incorporating a proximal term. FAVOR \cite{wang2020favor} ameliorates the bias induced by non-IID data by selecting a subset of participating clients in each training round. FedMA \cite{wang2020fedma} introduces a layer-wise approach that leverages Bayesian non-parametric methods to mitigate data heterogeneity. FedGen \cite{venkateswaran2023fedgen} employs a masking function to address spurious correlations and biases in the training data, enabling clients to identify and differentiate between spurious and invariant features. MOON \cite{li2021moon} capitalizes on the similarity between models to refine the training process of individual clients, achieving superior performance in federated learning for image domains. FedNTD \cite{lee2022fedntd} utilizes Knowledge Distillation to alleviate the issue of data heterogeneity, concentrating solely on data that has not been accurately predicted.

\subsection{Personalized Federated Learning}
Recently, personalized federated learning (pFL) has attracted significant attention within the research community due to its superior performance in addressing data heterogeneity challenges \cite{kairouz2021advances}. In the Ditto framework \cite{li2021ditto}, each client incorporates an optimal term to extract information from the global model, learning an additional personalized model. FedBABU \cite{oh2021fedbabu} fine-tunes the classifier within the global model using client-specific data to develop personalized models for each client. FedALA \cite{zhang2023fedala} introduces an adaptive aggregation strategy to selectively assimilate information from the global model. 
\section{Methodology}
\label{se:Meth}
\subsection{Problem Statement}
Federated Learning aims to train a global model on the server while ensuring the data privacy of clients. Suppose there are $N$  clients, with the data of the $i$-th client denoted as $D^i$. Let $\mathcal{L}(\cdot,\cdot)$ represent the loss function of each local model. Typically, we minimize the $Loss$, as defined in Equation \eqref{eq:loss}, to obtain the global model $f(\cdot)$.

\begin{equation}
    Loss=\frac{\sum_{i=1}^N|D^i|\cdot\mathbb{E}_{(X^i,Y^i)\sim D^i}[\mathcal{L}(f(X^i),Y^i)]}{\sum_{i=1}^N|D^i|}.
    \label{eq:loss}
\end{equation}

\subsection{Our Method}
\label{se:method}
In our experiments, we observe that the non-IID issue exacerbates the dispersion among local models due to the varying information available to different clients. Conversely, under the IID hypothesis, the dispersion is significantly reduced, as shown in Figure \ref{fig:motivation}
. Motivated by the aforementioned observations, we propose \textbf{Fed}erated \textbf{Pa}rameter S\textbf{ke}w Learning (FedPake), a novel and effective FL algorithm, shown in Figure \ref{fig:architecture}. FedPake aims to enhance the global model's robustness by reallocating the weight of each parameter according to the extent of \textit{parameter skew}.

\begin{algorithm}[ht]
    \caption{FedPake}
    \label{algo}
        \textbf{Input} :
        $\mathcal{K}$: client collection, 
        $\rho$: client joining ratio, 
        $\Theta^{0}$: initial global model, 
        $C$: number of Micro-Class,
        $S$: number of Macro-Class, 
        $\lambda$: the threshold between high-dispersion and low-dispersion, 
        $T$: train round. \\
        \textbf{Output}: 
        Finial global model $\hat{\Theta}^T$.
    \begin{algorithmic}[1]
        \State Sever sends $\Theta^{0}$ to all clients to initialize local models.
        \For{$each \ round \ t = 1,\cdots, T$}
            \State Sample clients $\mathcal{K}^t \leftarrow \operatorname{Sample}(\mathcal{K}, \rho)$.
            \For{$each \ client \ k\in \mathcal{K}^t$ \textbf{in parallel}}
                \State Client update local model $\hat{\Theta}^{t-1}_{k} \leftarrow \hat{\Theta}^{t-1}$.
                \State Client train local model $\hat{\Theta}^{t}_{k} \leftarrow \operatorname{Train}(\hat{\Theta}^{t-1}_{k})$.
            \EndFor
            \State Server collects local models $\hat{\Theta}^{t}_{\mathcal{K}^t} = \{ \hat{\Theta}^{t}_k \}, k\in \mathcal{K}^t$.
            \State \textcolor[RGB]{0,156,156}{$\# Server \ aggregates \ local \ models.$}
            \For{$each \ layer \ l=1,\cdots, L$}
                \State $\mathcal{W} \leftarrow \Theta^{t}_{\mathcal{S}^t , l}$
                \State $\mathbf{r}^h, \mathbf{r}^l \leftarrow \operatorname{ParameterDivision}(\mathbf{w})$.
                \State \textcolor[RGB]{0,156,156}{$\# Micro-Class.$}
                \State Categorize $\mathcal{W}$ into Micro-Class,
                \State $\mathbf{E} \leftarrow$ Equation\eqref{eq:06}.
                \State \textcolor[RGB]{0,156,156}{$\# Macro-Class.$}
                \State Categorize clients into Macro-Class,
                \State $\{ \mathcal{G}_1^t, \ldots, \mathcal{G}_S^t \} \leftarrow \operatorname{ClusterAnalysis}(\mathbf{E})$,
                \State \textcolor[RGB]{0,156,156}{$\# Compute \ the \ aggregation \ weight.$}
                \For{$each \ Macro-Class \ j=1, \ldots, S$}
                    \State $\mathbf{Q}_j \leftarrow$ Equation\eqref{eq:08}.
                    \State $\alpha_j \leftarrow$ Equation\eqref{eq:10}.
                \EndFor
                \State $\hat{\Theta}^t_l \leftarrow \hat{\mathcal{W}} \leftarrow$ Equation\eqref{eq:09}.
            \EndFor
            \State \textcolor[RGB]{0,156,156}{$\# Update \ global \ model \ parameters.$}
            \State $\hat{\Theta}^t = \{ \hat{\Theta}^t_1,\ldots,\hat{\Theta}^t_L \}$.
        \EndFor
        \\
        \Return  $\hat{\Theta}^{T}$
    \end{algorithmic}
\end{algorithm}

By calculating the dispersion of parameters from different clients, FedPake divides the parameters into high-dispersion and low-dispersion using the threshold $\lambda$. For the low-dispersion parameter, we calculate the client-dimension average of parameters, while for the high-dispersion parameter, we compute weight $\mathbf{\alpha}$ to aggregate global model parameters based on the Micro-Class distribution in the Macro-Class. Details are illustrated in Algorithm \ref{algo}. Here, we set client collection as $\mathcal{K}=\{ k_1, k_2, \ldots, k_N \}$;
Each client model includes $L$ layers, and we demonstrate our methodology using $l$-th layer $\mathbf{w}$ as an example, where the number of parameters is denoted by $M$,  namely all client model parameters denote as $\mathcal{W}=\{\mathbf{w}_{k_1}, \dots, \mathbf{w}_{k_N}\} \in \mathbb{R}^{|\mathcal{K}| \times M}$.

\noindent \textbf{Parameter Division.}
\label{se:RSR}
 We use the coefficient of variation ($\mathbf{cv}$) \cite{brown1998coefficient}, which owns prominent ability to discriminate dispersion statistically, to measure the discrepancies among the parameters from disparate clients.

\begin{equation}
    \mathbf{cv} = \frac{[\operatorname{Mean}\big((\mathcal{W} - \bar{\mathcal{W}})^{(2)}\big)]^{\frac{1}{2}}}{\bar{\mathcal{W}}} \in \mathbb{R}^{1 \times M},
    \label{eq:03}
\end{equation}
\begin{equation}
    \bar{\mathcal{W}} = \operatorname{Mean}(\mathcal{W}) \in \mathbb{R}^{1 \times M},
\end{equation}
where $\operatorname{Mean}(\cdot)$ represents that the first dimension is averaged. Based on threshold $\lambda$, we obtain the high-dispersion region  $\mathbf{r}^h$  and the low-dispersion region $\mathbf{r}^l$, and $\mathbb{I}(\cdot)$ is indicator function:

\begin{equation}
    \mathbf{r}^h = \mathbb{I}(\frac{
    \mathbf{cv} - min(\mathbf{cv})
    }{
    max(\mathbf{cv}) - min(\mathbf{cv})
    } > \lambda) \in \mathbb{R}^{1 \times M}, 
    \label{eq:04}
\end{equation}

\begin{equation}
    \mathbf{r}^l = \mathbb{I}(\frac{
    \mathbf{cv} - min(\mathbf{cv})
    }{
    max(\mathbf{cv}) - min(\mathbf{cv})
    } \leq \lambda) \in \mathbb{R}^{1 \times M}.
    \label{eq:05}
\end{equation}

In $\mathbf{r}^l$, the variation among the parameters of the clients' models is under $\lambda$. Consequently, we compute the average of these parameter values to determine the parameters of the global model.

\noindent \textbf{Micro-Class.}
\label{se:MIC}
For the parameters exhibiting significant differences among clients, denoted as $ \mathbf{r}^h$, these are the areas of primary focus. The variation in these parameters reflects, to a certain extent, the distinct characteristics of the clients' models.

Due to the limitations of using a fixed threshold, which does not adequately account for the extent of parameter discrepancies in the high-dispersion region, we introduce Micro-Class and Macro-Class to describe the discrepancies. Micro-Class and Macro-Class, respectively, assess dispersion from the perspectives of local parameters and the global network. 

Micro-Class is formulated as follows:

\begin{equation}
    \mathbf{E} = \sum_{i=1}^C i \cdot \mathbb{I}
    (
    \frac{i}{C} \geq 
    (\mathcal{W} - \bar{\mathcal{W}})^{(2)} >
    \frac{i-1}{C}
    ) 
    \in \mathbb{R}^{|\mathcal{K}| \times M}
    \label{eq:06},
\end{equation}
\begin{equation}
    \mathbf{E} = \{E_{k_{1}},\dots,E_{k_{N}}\},  E_{k_{n}} \in \mathbb{R}^{1 \times M},
\end{equation}
where $C$ is the number of Micro-Class, $E_{k_{n}}$ records the distribution of Micro-Class in $k_n$ client model parameters. Equation\eqref{eq:06} has the following properties: (1)$\bigcap\limits_{i=1}^{C} \mathbb{I}(\frac{i}{C} \geq (\mathcal{W} - \bar{\mathcal{W}})^{(2)} > \frac{i-1}{C}) = \emptyset$; (2)$\bigcup\limits_{i=1}^{C} \mathbb{I}(\frac{i}{C} \geq (\mathcal{W} - \bar{\mathcal{W}})^{(2)} > \frac{i-1}{C}) = \mathbf{1}_{|\mathcal{K}| \times M}$.

\noindent \textbf{Macro-Class.}
\label{se:MAC}
Intuitively, the training of a model is a holistic process wherein local models exhibit synergistic effects. Micro-Class only considers the differences between clients in a local parameter perspective. However, the differences in the global synergistic effects among local models are also crucial aspects, which can better reflect the characteristics of the models. Therefore, we propose Macro-Class, which measures the similarity between clients based on the distribution of Micro-Class across each model. We cluster and map the clients according to their Micro-Class similarity, resulting in Macro-Class, which effectively captures the exclusive information of each class of client.

Directly measuring the similarity between clients is challenging. Therefore, we derive the similarity by calculating the degree of dissimilarity between clients:
\begin{equation}
    \operatorname{SIM}(k_1,k_2) = 
    1 - \frac{
    \operatorname{Count}(
    \mathbb{I}(E_{k_1} \neq  E_{k_2})
    )
    }{
    \operatorname{Count}(
    \mathbf{r}^h
    )
    }.
    \label{eq:07}
\end{equation}

\begin{table*}[ht]
  \centering
  \resizebox{0.95\linewidth}{!}{
    \begin{tabular}{l|*{3}{c}|*{4}{c}}
    \toprule
    Settings & \multicolumn{3}{c|}{Pathological heterogeneous setting} & \multicolumn{4}{c}{Practical heterogeneous setting} \\
    \midrule
    Methods & CIFAR-10 & CIFAR-100 & Tiny-ImageNet & CIFAR-10 & CIFAR-100 & Tiny-ImageNet & CIFAR-100*  \\
    \midrule
    FedAvg & 90.79$\pm$0.08 & 50.19$\pm$0.31 & 33.58$\pm$0.15 & 88.55$\pm$0.10 & 33.57$\pm$0.09 & 19.86$\pm$0.20 & 34.39$\pm$0.31 \\
    FedProx & 90.75$\pm$0.08 & 50.08$\pm$0.30 & 32.98$\pm$0.08 & 88.94$\pm$0.08 & 34.10$\pm$0.39 & 19.64$\pm$0.22 & 34.39$\pm$0.30 \\
    MOON & 90.65$\pm$0.16 & 50.42$\pm$0.11 & 33.82$\pm$0.07 & 88.78$\pm$0.25 & 33.91$\pm$0.15 & 19.72$\pm$0.15 & 34.64$\pm$0.04 \\
    FedGEN & 90.52$\pm$0.10 & 50.38$\pm$0.66 & 32.77$\pm$0.42 & 88.84$\pm$0.23 & 34.16$\pm$0.17 & 19.42$\pm$0.50 & 35.00$\pm$0.11 \\
    FedNTD & 90.22$\pm$0.12 & 50.71$\pm$0.49 & 34.05$\pm$0.47 & 88.60$\pm$0.10 & 33.90$\pm$0.32 & 19.57$\pm$0.09 & 34.79$\pm$0.45 \\
    \midrule
    Ditto & 90.53$\pm$0.04 & 50.27$\pm$0.35 & 33.27$\pm$0.23 & 88.87$\pm$0.23 & 34.05$\pm$0.19 & 19.84$\pm$0.37  & 34.55$\pm$0.22 \\
    FedBABU & 90.02$\pm$0.14 & 64.86$\pm$0.24 & 36.09$\pm$0.52 & 88.20$\pm$0.27 & 36.01$\pm$0.36 & 22.02$\pm$0.31 & 37.15$\pm$0.24 \\
    FedALA & 90.47$\pm$0.28 & 50.10$\pm$0.28 & 33.26$\pm$0.25 & 88.81$\pm$0.15 & 33.75$\pm$0.04 & 19.62$\pm$0.22 & 34.80$\pm$0.07 \\
    \midrule
    FedPake(ours) & \textbf{91.36$\pm$0.23} & \textbf{69.72$\pm$0.25} & \textbf{39.27$\pm$0.39} & \textbf{90.41$\pm$0.17} & \textbf{39.09$\pm$0.19} & \textbf{25.41$\pm$0.37} & \textbf{39.12$\pm$0.04} \\
    \bottomrule
    \end{tabular}}
  \caption{\textbf{The test accuracy (\%) in the pathological heterogeneous setting and practical heterogeneous setting.}}
    \label{tab:comparison}
\end{table*}

Here, $\operatorname{Count}(\cdot)$ returns the number of non-zero elements. $\operatorname{SIM}(k_1,k_2)$ represents the Micro-Class similarity between client $k_1$ and client $k_2$. We cluster the clients according to their Micro-Class similarity derived from Equation \eqref{eq:07}. And we employ the hyperparameter $S$ to limit the maximum number of clusters.

The clusters are denoted as $\{ \mathcal{G}_1, \mathcal{G}_2, \ldots, \mathcal{G}_S \}$, where $\mathcal{K} = \bigcup\limits_{j=1}^{S} \mathcal{G}_j$ and $\mathcal{G}_j$ records the clients contained in the $j$-th cluster, namely the cluster is the Macro-Class. Macro-Class encapsulates the tendencies of clients, which we refer to as $\mathbf{Q}$. $\mathbf{Q}_{j}$ is derived from the Micro-Class mapping of clients within the $j$-th Macro-Class:
\begin{equation}
    \mathbf{Q}_{j} = \mathbf{r}^h \odot
    \operatorname{Top} \left( 
    \mathbf{E}_{\mathcal{G}_j}
    \right )
    \in \mathbb{R}^{1 \times M},
    \label{eq:08}
\end{equation}
where $\mathbf{E}_{\mathcal{G}_j}$ denotes all $E$ for clients within the $\mathcal{G}_j$. $\operatorname{Top}(\cdot)$ returns the Micro-Class with the highest frequency along the first dimension.

\noindent \textbf{Aggregation.}
In this section, we introduce a novel strategy for updating the global model, enhancing the model's quality. The parameters of the global model, denoted as $\hat{w}$, are comprised of low-dispersion parameters and high-dispersion parameters. The procedure for computing these parameters is outlined as follows:
\begin{equation}
    \hat{w} = 
    \mathbf{r}^l \odot \frac{1}{|\mathcal{K}|} \sum_{k \in \mathcal{K}} \mathbf{w}_k +
    \mathbf{r}^h \odot \sum_{j=1}^{S} \alpha_j (\frac{1}{|\mathcal{G}_j|} \sum_{k \in \mathcal{G}_j} \mathbf{w}_k),
    \label{eq:09}
\end{equation}
\begin{equation}
    \mathbf{\alpha}_j =
    \sum_{i=1}^{C}
    \frac{
    \operatorname{Count}(
    \mathbb{I} ( \mathbf{Q}_{j} = i )
    )
    }{
    S \times \operatorname{Count}(\mathbf{r}^h)
    } \cdot
    \mathbb{I} ( \mathbf{Q}_{j} = i )
    \in \mathbb{R}^{1 \times M}.
    \label{eq:10}
\end{equation}
 We compute aggregation weight $\mathbf{\alpha}_j$ for client model parameter in $j$-th Macro-Class based on the distribution of Micro-Class. And property $\bigcap\limits_{i=1}^{C} \mathbb{I} ( \mathbf{Q}_{j} = i ) = \emptyset$ guarantees that the weights will not be repeated. A detailed instance is presented in Appendix Figure 7.
\section{Experiments}
\subsection{Experiment Setting}

\noindent \textbf{Baselines.}
In this section, we select eight Federated Learning methods, encompassing both traditional FL (tFL) and personalized FL (pFL). TFL includes FedAVG \cite{mcmahan2017communication}, FedProx \cite{li2020fedPROX}, MOON \cite{li2021moon}, FedNTD \cite{lee2022fedntd}, Scaffold\cite{karimireddy2020scaffold}, and FedDyn\cite{feddyn}. Given that FedPake focuses on distinctive global models, to comprehensively analyze our model's performance, we also choose SOTA pFL methods that are capable of generating global models, including Ditto \cite{li2021ditto}, FedBABU \cite{oh2021fedbabu}, FedALA \cite{zhang2023fedala}, and FedConcat\cite{fedconcat}. Additionally, we illustrate the advantages of FedPake by utilizing the distribution of local models' parameters while also demonstrating the reason why other FL methods are unable to achieve good performance.

\begin{figure*}
    \centering
    \includegraphics[width=1\linewidth]{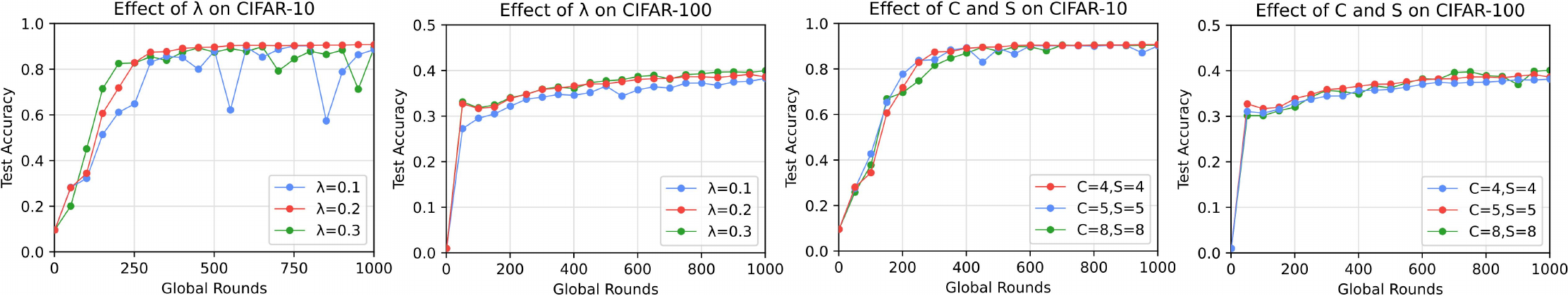}
    \caption{\textbf{The effectiveness of each hyperparameter.} On CIFAR-10/100, we demonstrate the training of FedPake with various hyperparameter values, including $\lambda$, $C$, and $S$. And others follow the default experiment setting. {\color{red} Red line} is the optimal hyperparameter setting. }
    \label{fig:fine}
\end{figure*}

\noindent \textbf{Datasets.}
Our experiments are conducted on three widely used Computer Vision datasets, including CIFAR-10/100 and Tiny-ImageNet. We also give an introduction for three datasets in Appendix A.1. The ratio of the train set to the test set is 0.75: 0.25. In this work, we simulate heterogeneous settings using pathological heterogeneity \cite{mcmahan2017communication,shamsian2021personalized,zhang2023fedala} and practical heterogeneity \cite{lin2020ensemble,li2021model}.
Regarding pathological heterogeneity, we allocate 2, 10, and 20 classes for CIFAR-10, CIFAR-100, and Tiny-ImageNet, respectively, to each client, while the classes for each client in the practical heterogeneity are controlled by the Dirichlet distribution $\operatorname{Dir(\beta)}$. The smaller the $\beta$, the more severe the data heterogeneity. In this work, we set $\beta$=0.1 for the experiments. Moreover, in Appendix A.2, we present more details about heterogeneity data.

\noindent \textbf{Train Setting.}
To tackle the limitations of the simple CNN backbone in demonstrating the performance of FedPake, we employ ResNet-18 \cite{he2016deep} as our backbone. Additionally, CIFAR-100* represents that we carry out experiments with ResNet-34 \cite{he2016deep} backbone on the CIFAR-100 dataset. On the server side, we set the global training rounds to 1000, the number of clients to 20, and the proportion of randomly selected clients per round $\rho$ to 1.0. On the client side, we set the local training rounds per client to 1. Additionally, we compute the average testing accuracy of the global model over the last 10 rounds. Additionally, we run all tasks three times and report the mean and standard deviation in the Table\ref{tab:comparison}, Table \ref{tab:heterogeneity}, and Table \ref{tab:scalability}. Details of settings of different methods are shown in Appendix C.

\noindent\textbf{Hyperparameter Setting.}
Hyperparameter settings determine the performance of our method, so we set the optimal hyperparameters in this section. Additionally, details about how to choose the values of them are shown in the later analysis. We set the threshold $\lambda$ for distinguishing high-dispersion and low-dispersion regions to 0.2. Specifically, for CIFAR-10/100 and Tiny-ImageNet, we set the number of Micro-Class $C$ to 4, 5, and 8, and the number of Macro-Class $S$ to 4, 5, and 8, respectively.
\begin{table*}[ht]
  \centering
  \resizebox{0.85\linewidth}{!}{
    \begin{tabular}{l|*{3}{c}|*{3}{c}}
    \toprule
     & \multicolumn{6}{c}{Scalability}  \\
    \midrule
    Datasets & \multicolumn{3}{c|}{CIFAR-10} & \multicolumn{3}{c}{CIFAR-100} \\
    \midrule
     \multirow{2}{*}{Methods} & $\rho = 1.0$ & \multicolumn{2}{c|}{$\rho = 0.5$} & $\rho = 1.0$ & \multicolumn{2}{c}{$\rho = 0.5$} \\
     & 50 clients & 50 clients & 100 clients & 50 clients & 50 clients & 100 clients \\
    \midrule
    FedAvg & 84.80$\pm$0.24 & 84.91$\pm$0.31 & 88.09$\pm$0.19 & 29.71$\pm$0.11 & 29.72$\pm$0.16 & 30.64$\pm$0.19\\
    FedProx & 84.86$\pm$0.27 & 84.24$\pm$0.66 & 88.45$\pm$0.13 & 30.18$\pm$0.51 & 30.14$\pm$0.09 & 30.49$\pm$0.39\\
    MOON & 84.64$\pm$0.18 & 84.47$\pm$0.59 & 87.72$\pm$0.18 & 29.91$\pm$0.24 & 29.59$\pm$0.18 & 30.65$\pm$0.33 \\
    FedGen & 84.73$\pm$0.31 & \textbf{85.65$\pm$0.30} & 87.67$\pm$0.38 & 29.74$\pm$0.09 & 30.64$\pm$0.21 & 30.36$\pm$0.59 \\
    FedNTD & 83.40$\pm$0.10 & 83.95$\pm$0.09 & 86.33$\pm$0.27 & 29.58$\pm$0.19 & 29.90$\pm$0.24 & 30.47$\pm$0.19 \\
    \midrule
    Ditto & 84.34$\pm$0.22 & 84.91$\pm$0.54 & 88.02$\pm$0.20 & 29.44$\pm$0.21 & 30.24$\pm$0.24 & 30.39$\pm$0.13 \\
    FedBABU & 84.06$\pm$0.10 & 84.57$\pm$0.19 & 87.02$\pm$0.19 & 30.08$\pm$0.28 & 30.11$\pm$0.45 & 30.79$\pm$0.30 \\
    FedALA & 84.18$\pm$0.26 & 84.76$\pm$0.21 & 87.81$\pm$0.14 & 29.60$\pm$0.29 & 29.61$\pm$0.27 & 30.44$\pm$0.07 \\
    \midrule
    FedPake(ours) & \textbf{87.24$\pm$0.14} & 84.53$\pm$0.13 & \textbf{88.63$\pm$0.22} & \textbf{31.83$\pm$0.35} & \textbf{31.31$\pm$0.20} & \textbf{31.10$\pm$0.38} \\
    \bottomrule
    \end{tabular}}
  \caption{\textbf{The test accuracy (\%) with various the number of clients and client joining ratio $\rho$ on CIFAR-10/100}. And, except for the number of clients and $\rho$, others are set to the default.}
    \label{tab:scalability}
\end{table*}

\begin{figure}
    \centering
    \includegraphics[width=0.7\linewidth]{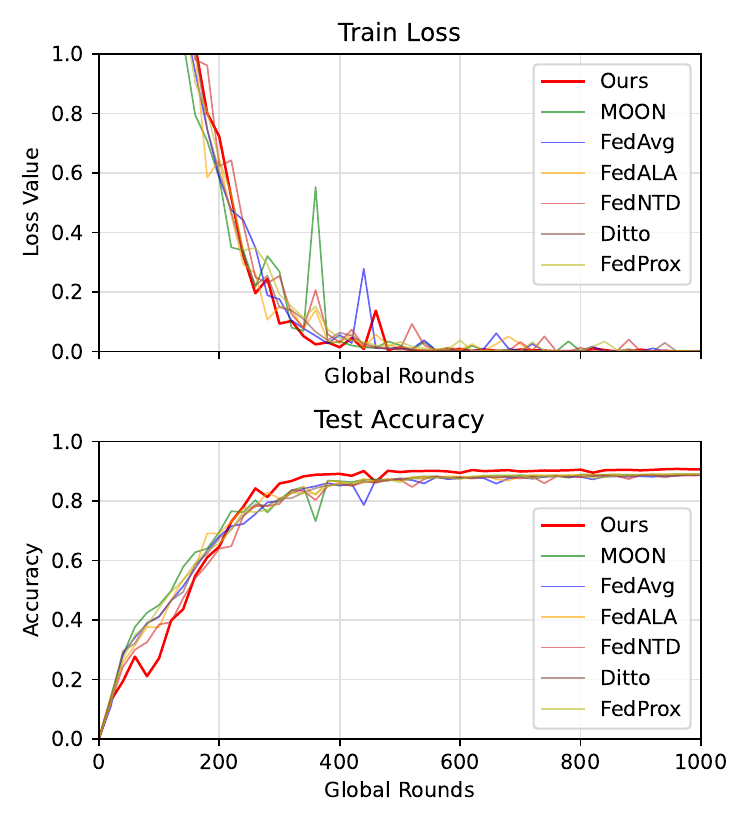}
    \caption{\textbf{The performance of FedPake(ours) and baselines on CIFAR-10.} The \textbf{top} figure presents the training loss of seven FL methods, and the \textbf{bottom} shows their test accuracy. Experiments are conducted under default settings.}
    \label{fig:loss}
\end{figure}

\subsection{Performance Comparison and Analysis}
\noindent \textbf{Results.}
We evaluate the performance of FedPake against other baselines under different settings. The results in Table \ref{tab:comparison} illustrate that FedPake consistently outperforms the other eight FL algorithms in all scenarios. Specifically, in challenging CIFAR-100, our method offers an average improvement of about 5\% and 3\%, respectively. When considering a pathological heterogeneous setting, where label classes from each dataset are fixed, notable improvements in accuracy are evident despite label skew. While the label skew is slight in practical heterogeneous settings, due to the Dirichlet-based label distribution of datasets, FedPake outperforms others by about 3\%. Meanwhile, FedBABU, a classical algorithm in pFL, trains professional local models with fine-tuning for each client. Our method not only surpasses FedBABU in every scenario but also has lower computation costs. Due to the poor generalization ability of the global model, FedAVG and FedProx perform poorly in two settings. Moreover, we conduct experiments on more recent methods (FedConcat) and other classic methods (FedDyn and Scaffold) on CIFAIR10, as shown in Appendix Table 5. Our method surpasses FedConcat and Scaffold by 2.2\% on average. FedDyn's performance is higher than FedPake by only about 0.5\%, but our method's computational expense is lower than FedDyn by about 7.8\%, which suggests that our method is more efficient. These results further show FedPake's powerful capability.

In Figure \ref{fig:loss}, we show the evolution of train loss and test accuracy on CIFAR-10 from our method and all baselines. The figure shows that FedPake converges faster and achieves the highest accuracy compared with other baselines. After 500 rounds, the loss convergence curve of FedPake becomes quites stable, while the loss of other methods continues to oscillate, highlighting FedPake’s smooth convergence.

\noindent \textbf{Effect of hyperparameter.}
To make our method more practical, we conducted experiments under different hyperparameter settings on CIFAR-10/100 to illustrate the effect of hyperparameter, the results are shown in Figure \ref{fig:fine}. {\color{red}Red line} is the test accuracy of the model with the optimal hyperparameter setting. A larger $\lambda$ means the high-dispersion region is smaller. As shown in the figure, FedPake converges faster as $\lambda$ gets larger, and $\lambda$ also impacts the stability of model training. To balance the convergence rate and stability, we set the best value, $\lambda = 0.2$, in our experiments. $C$ and $S$ represent the maximum number of Micro-Class and Macro-Class, respectively. Recall from Section \ref{se:Meth}, since larger $C$ and $S$ mean higher computation cost of FedPake, we do extensive experiments to find the most outstanding performance of our method with relatively small $C$ and $S$. On CIFAR-10, experiments show our model could achieve the best test accuracy with the minimal computation expense, i.e, $C = 4$ and $S=4$. Furthermore, when $S=8$ and $C=8$, the performance of FedPake on CIFAR-100 is the best. However, comparing $S=5$ and $C=5$, the performance doesn't present a significant increase under $S=8$ and $C=8$, which also raises computing expenses. Therefore, we set $S=5$ and $C=5$ for experiments on CIFAR-100 while $C = 4$ and $S=4$ on CIFAR-10. Conclusively, following our hyperparameter recommendation, our method could perform excellently in the real world.

\begin{table}[ht]
  \centering
  \resizebox{\linewidth}{!}{
    \begin{tabular}{l|*{3}{c}}
    \toprule
    & \multicolumn{3}{c}{Heterogeneity}  \\
    \midrule
    Methods & $\operatorname{Dir(0.01)}$ & $\operatorname{Dir(0.5)}$ & $\operatorname{Dir(1)}$ \\
    \midrule
    FedAvg & 48.42$\pm$0.48 & 36.04$\pm$0.18 & 37.46$\pm$0.14 \\
    FedProx & 48.62$\pm$0.66 & 36.39$\pm$0.19 & 37.47$\pm$0.12 \\
    MOON & 47.76$\pm$0.41 & 36.07$\pm$0.18 & 37.21$\pm$0.11 \\
    FedGen & 51.07$\pm$0.81 & 36.18$\pm$0.28 & 37.61$\pm$0.22 \\
    FedNTD & 48.48$\pm$0.71 & 36.38$\pm$0.35 & 37.63$\pm$0.10 \\
    \midrule
    Ditto & 48.57$\pm$0.71 & 36.18$\pm$0.26 & 37.55$\pm$0.22 \\
    FedBABU & 71.67$\pm$0.16 & 35.44$\pm$0.12 & 36.58$\pm$0.42 \\
    FedALA & 48.34$\pm$0.30 & 36.32$\pm$0.15 & 37.37$\pm$0.03 \\
    \midrule
    FedPake(ours) & \textbf{74.39$\pm$0.66} & \textbf{38.13$\pm$0.41} & \textbf{39.41$\pm$0.10}\\
    \bottomrule
    \end{tabular}}
  \caption{\textbf{The test accuracy (\%) with different heterogeneous settings on  CIFAR-100}. In addition to heterogeneous settings $\beta$, others follow the default experiment setting. }
    \label{tab:heterogeneity}
\end{table}

\noindent \textbf{Heterogeneity.}
 To study the effectiveness of FedPake in settings with different degrees of heterogeneity, we vary the $\beta$ in $\operatorname{Dir(\beta)}$ on CIFAR-100. The smaller $\beta$ is, the more heterogeneity the setting is. In Table \ref{tab:heterogeneity}, when heterogeneity is highest ($\beta$ = 0.01), our model outperforms eight baselines with an average improvement of 45.12\%. As heterogeneity increases (with $\beta$ decreasing from 1.0 to 0.01), FedPake’s performance on CIFAR-100 improves significantly (from 39.41\% to 74.39\%). This indicates that FedPake has excellent adaptability in high-heterogeneity environments.

\noindent \textbf{Scalability.} 
To demonstrate the scalability of FedPake, we conduct comprehensive experiments with 50 and 100 clients, using $\beta = 0.1$ in the practical heterogeneous setting. In Table \ref{tab:scalability}, most FL methods experience significant degradation when the number of clients increases and adopt a different client joining ratio $\rho$. On the CIFAR-100 dataset, FedPake achieves a test accuracy that surpasses the state-of-the-art (SOTA) by 1.65\% when $\rho = 1.0$, but the improvement is reduced to only 0.67\% when $\rho = 0.5$. This performance drop can be attributed to the fact that when $\rho=0.5$, only 50\% of clients participate in model training, limiting the amount of information available to FedPake and thereby diminishing its performance. On the CIFAR-10 dataset, when all clients are involved in training, namely $\rho = 1.0$, our method's performance improves from 84.53\% to 87.24\%, whereas the performance of other methods remains relatively unchanged. These results demonstrate that our model's outstanding scalability compared with other baselines, which highlight FedPake's applicability in the real world.

\noindent \textbf{Communication and Computation Cost.   }
We record the total time cost for each method until convergence, as shown in Appendix Table 6. FedPake costs 0.248 min (similar to FedAVG) in each iteration. In other words, our method only costs an additional 0.002 min for great accuracy improvement. Moreover, we show the communication cost for one client in one iteration in Appendix Table 6. The communication overhead for most methods is the same as FedAVG, which uploads and downloads only one model.
\begin{table}[htp]
\centering
\resizebox{0.4\textwidth}{!}{
    \begin{tabular}{ccccc} 
    \toprule 
    Micro-Class & 
    Macro-Class &
    CIFAIR-10\\
    \midrule
     $-$ & $-$ & 24.51  \\
     $-$ & \checkmark & 14.86  \\
     \checkmark & $-$  & 17.74  \\
     \checkmark & \checkmark & \textbf{90.60}  \\
    \bottomrule
    \end{tabular}
    }
    \caption{\textbf{Ablation Study about Micro-Class and Macro-Class on CIFAIR-10.}}
    \label{tab:ablation}
\end{table}

\subsection{Model Analysis}
\noindent \textbf{Ablation Study.}
We conduct ablation studies on Macro-Class and Micro-Class, as shown in Table \ref{tab:ablation}. Since Macro-Class and Micro-Class are integral to FedPake, removing either component prevents the model from functioning. In the ablation study, we randomly initialize either Macro-Class or Micro-Class as w/o. In Table \ref{tab:ablation}, removing Macro-Class or Micro-Class drastically reduces accuracy from 90.60\% to as low as 14.86\% or 17.74\%. This sharp drop demonstrates the synergy between the coarse-grained grouping (Macro-Class) and the fine-grained distinctions (Micro-Class), confirming that both properties are critical for the model’s performance.

\begin{figure}
    \centering
    \includegraphics[width=1\linewidth]{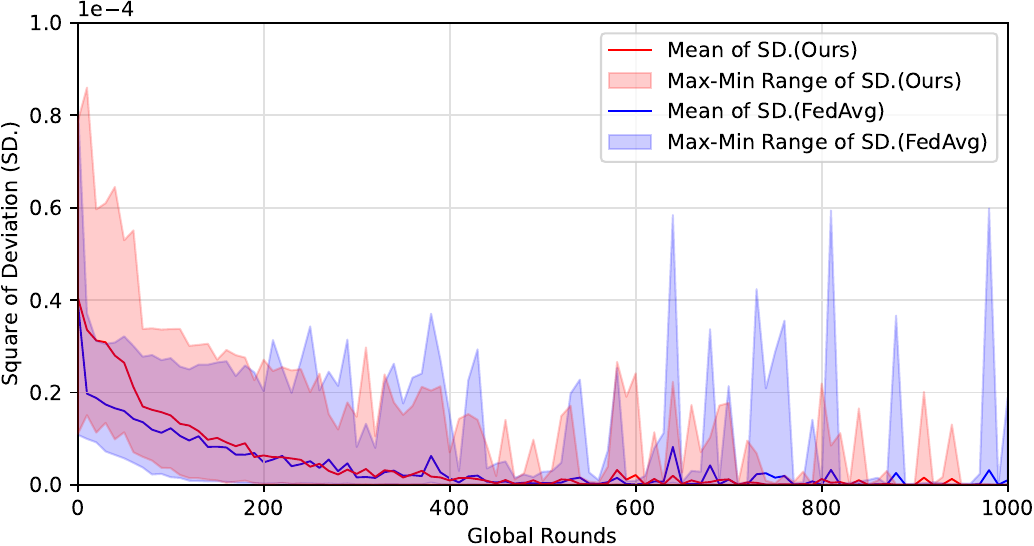}
    \caption{\textbf{The parameter value dispersion of each client model.} Under CIFAR-10, we illustrate the change of parameter value dispersion on {\color{red} FedPake} and {\color{blue} FedAVG} during training. Additionally, the square of deviation (SD) represents the extent of dispersion. Hyperparameter settings follow the default.}
    \label{fig:showsd}
\end{figure}

\noindent \textbf{Parameter Skew Analysis.}
\textit{Parameter skew} can undermine the stability of the FL algorithm when the parameters of the global model are aggregated, with extreme values in the client model parameters being a primary cause of this skew. In Figure \ref{fig:showsd}, we illustrate the evolution of the dispersion in client model parameter values throughout training. As observed, after 600 rounds of training with FedPake, both the mean and range of the squared deviation (SD) are significantly reduced, indicating a substantial decrease in the occurrence of extreme parameter values in the client models. So, our method outperforms FedAVG in addressing the issue of extreme parameter values.

To further understand our method's effectiveness, we show the value distribution of parameters in local models and global models from different FL methods in Figure \ref{fig:analysis}. The distribution of most parameters across different client models is skewed, which supports \textit{parameter skew}. This skewness contributes to the weak generalization ability of the FL algorithm due to the unrobust estimation of the global model. For example, FedAVG directly averages these parameters, but according to the Law of Large Numbers \cite{hsu1947complete}, the mean of a skewed distribution can be biased by extreme values, resulting in overestimation or underestimation, which can hinder the global model's robustness and generalizability.

Detailly, as shown in Figure \ref{fig:analysis}, the parameters of the FedAVG global model deviate from the main peak of distribution. However, by accounting for \textit{parameter skew}, FedPake adjusts the global model's parameter values to better align with the overall trend, resulting in relatively unbiased estimation values. Therefore, our method enables the global model to capture the characteristics of most local models, enhancing its generalizability. This explains the significant advantage of FedPake in updating the global model's parameters.

\section{Conclusion and Discussions}

\subsection{Conclusion}
Federated learning (FL) has become a promising method to resolve the pain of silos in many domains such as medical imaging and micro-model deployment. The heterogeneity of data is the key challenge for the performance of FL. We propose FedPake, a novel and conductive approach for FL, to enhance the performance of federated deep learning models on non-IID datasets. FedPake introduces a conception, \textit{parameter skew}, and tackles the implication of it. Our extensive experiments show that FedPake achieves great improvements over SOTA approaches in various scenarios. Moreover, we analyze the effectiveness of our method and demonstrate why other FL methods fail to achieve good performance.

\subsection{Discussions}
\noindent \textbf{Limitation.}
For lack of computation resources and a tight schedule, at this time, we could not further compare many FL methods and fully investigate the potential of FedPake on larger models.

\noindent \textbf{Future Work.}
FedPake is designed based on the discrete distance between clients, and we will further explore the performance of other methods for measuring discrepancy. Moreover, we will investigate the impact of \textit{parameter skew} on the Large Language Model and observe its performance under other settings.

{
    \small
    \bibliographystyle{ieeenat_fullname}
    \bibliography{main}
}

\appendix
\clearpage
\setcounter{page}{1}
\maketitlesupplementary

\section{Dataset}
\begin{table*}[htp]
    \centering
    \begin{tabular}{cccccccccccc} 
    \cline{1-12}
         \multicolumn{3}{c}{Method} & \multicolumn{3}{c}{Backbone} & \multicolumn{3}{c}{Accuracy} & \multicolumn{3}{c}{Total Time/ Iter Time} \\ 
    \cline{1-12}
         \multicolumn{3}{c}{FedConcat} & \multirow{2}{*}{\textbf{CNN}} & \multicolumn{2}{c}{} & \multicolumn{3}{c}{57.7} & \multicolumn{3}{c}{—} \\ 
         \multicolumn{3}{c}{FedPake (Ours)} &  & \multicolumn{2}{c}{} & \multicolumn{3}{c}{\textbf{59.7}} & \multicolumn{3}{c}{—} \\ 
    \cline{1-12}
         \multicolumn{3}{c}{Scaffold} & \multirow{3}{*}{\textbf{ResNet18}} & \multicolumn{2}{c}{} & \multicolumn{3}{c}{89.98$\pm$0.31} & \multicolumn{3}{c}{256min / 0.256min} \\ 
         \multicolumn{3}{c}{FedDyn} &  & \multicolumn{2}{c}{} & \multicolumn{3}{c}{\textbf{90.46}$\pm$0.06} & \multicolumn{3}{c}{264min / 0.267min} \\ 
         \multicolumn{3}{c}{FedPake (Ours)} &  & \multicolumn{2}{c}{} & \multicolumn{3}{c}{90.41$\pm$0.17} & \multicolumn{3}{c}{\textbf{225min / 0.248min}} \\ 
    \cline{1-12}
    \end{tabular}
    \caption{\textbf{The test accuracy (\%) and computational costs (Total Time\&Time / iter) in the practical heterogeneous setting on CIFAIR-10.} The backbone of FedConcat is a simple CNN, while the backbones of Scaffold and FedDyn are ResNet18.}
    \label{tab:adcompariment}
\end{table*}
\subsection{Information}
The experiments were conducted on CIFAR-10, CIFAR-100, and Tiny-ImageNet.
\begin{itemize}
    \item \textbf{CIFAR-10}~\cite{krizhevsky2009learning}
    A dataset published by CIFAR consists of 50,000 training images and 10,000 test images. It has 10 object classes, which include animals and vehicles, and the image size is $32 \times 32$.
    \item \textbf{CIFAR-100}~\cite{krizhevsky2009learning}
    It is an extended version of CIFAR-10 and consists of 50,000 training images and 10,000 test images. It has 100 object classes, each with 600 samples, and the image size is $32 \times 32$.
    \item \textbf{Tiny-ImageNet}~\cite{chrabaszcz2017downsampled}
    A dataset published by Stanford University consists of 100,000 training images, 10,000 validation images, and 10,000 test images. It has 200 object classes, each with 500 training samples, 50 validation samples, and 50 test samples, and the image size is $64 \times 64$.
\end{itemize}
For CIFAR-10/100, we merge the training and test images, then the training set and the test set are randomly divided according to the ratio of 0.75:0.25. For Tiny-ImageNet, since its test images are without labels, we merge the training and validation images. Therefore, CIFAR-10/100 dataset with 45,000 training images and 15,000 test images and Tiny-ImageNet dataset with 82,500 training images and 27,500 test images are used in our experiments.

\subsection{Heterogeneity Data}
We simulate heterogeneous settings using pathological heterogeneity and practical heterogeneity.
\begin{itemize}
    \item \textbf{Pathological Heterogeneity}~\cite{mcmahan2017communication,shamsian2021personalized,zhang2023fedala}
    A scenario only allows clients with a fixed number of labels.
    \item \textbf{Practical Heterogeneity}~\cite{lin2020ensemble,li2021model}
    A scenario assumes the label of client obey the Dilliclet distribution $\operatorname{Dir(\beta)}$ and sample images from dataset for each client base on distribution. Hyperparameter $\beta$ decides the degree of heterogeneity, and a higher value means data distribution closer to IID.
\end{itemize}
For Pathological Heterogeneity, since datasets have various numbers of labels, we allocate 2, 10, and 20 for CIFAR-10/100 and Tiny-ImageNet, respectively. For Practical Heterogeneity, we set $\beta = 0.1$ in default and $\beta = 0.01, 0.5, 1.0$ in the Heterogeneity experiment. Furthermore, Figure \ref{fig:label distribution(1)} shows the distribution of labels under the IID setting, pathological setting, and practical setting, and Figure \ref{fig:label distribution(2)} further shows the distribution of labels under different practical settings. 

\section{Experiments}
\subsection{Additional Experiments Result}
We conduct experiments on more recent methods (FedConcat) and other classic methods (FedDyn\cite{feddyn} and Scaffold\cite{karimireddy2020scaffold}) on CIFAIR10, as shown in Table \ref{tab:adcompariment}. Due to FedConcat's\cite{fedconcat} backbone being simple CNN, we also select simple CNN as the backbone to evaluate FedPake. Our method outperforms FedConcat by 3.8\%. Furthermore, in Table \ref{tab:adcompariment}, although FedDyn's performance is higher than FedPake by about 0.5\%, our method's computational expense is lower than FedDyn by about 7.8\%. It means that FedPake yields a significant computational cost reduction at the expense of a minimal loss of accuracy. FedPake surpasses Scaffold in both accuracy and computational costs. These results further suggest the outstanding capability of FedPake. Adabest\cite{varno2022adabest}, a classic FL method, is not open-source, so we can't conduct experiments on it.

\subsection{Communication and Computation Cost}
We record the total time cost for each method until convergence, as shown in Table \ref{tab:cost}. FedPake costs 0.248 min (similar to FedAVG) in each iteration. In other words, our method only costs an additional 0.002 min for great accuracy improvement. Moreover, we show the communication cost for one client in one iteration in Table \ref{tab:cost}. The communication overhead for most methods is the same as FedAVG, which uploads and downloads only one model.

\begin{table}[ht]
  \centering
  \resizebox{\linewidth}{!}{
    \begin{tabular}{l|rr|c}
    \toprule
    & \multicolumn{2}{c|}{Computation} & Communication\\
    \midrule
    Methods & Total time & Time/iter. & Param./iter. \\
    \midrule
    FedAvg & 208 min & 0.246 min & $2 * \Sigma$ \\
    FedProx & 245 min & 0.286 min & $2 * \Sigma$\\
    MOON & 402 min & 0.401 min & $2 * \Sigma$\\
    FedGen & 307 min & 0.508 min & $2 * \Sigma$\\
    FedNTD & 465 min & 0.302 min & $2 * \Sigma$\\
    \midrule
    Ditto & 265 min & 0.523 min & $2 * \Sigma$\\
    FedBABU & 246 min & 0.245 min & $2 * \alpha_f * \Sigma$\\
    FedALA & 210 min & 0.249 min & $2 * \Sigma$\\
    \midrule
    FedPake &  225 min & 0.248 min & $2 * \Sigma$\\
    \bottomrule
    \end{tabular}}
  \caption{\textbf{The computation cost on CIFAR-10 and the communication cost (transmitted parameters per iteration)}. $\Sigma$ is the parameter amount in the backbone. $\alpha_f$ ($\alpha_f < 1$) is the ratio of the parameters of the feature extractor in the backbone. }
    \label{tab:cost}
\end{table}

\section{Experimental Details}
\subsection{Hyperparameter Settings}
Since the results of baselines are reproduced, we use special instructions for the hyperparameter settings for all FL methods in this work.
\begin{itemize}
    \item For FedProx, the proximal term adjusts the distance between the local model and global model, we set the coefficient of proximal term $\mu$ to 0.001.
    \item For MOON, the temperature parameter controls the similarity between the local model and global model when calculating the contrast loss. we set the temperature parameter $\tau$ to 1.0.
    \item For FedGen, FedGen generates a generator and broadcasts it to all clients. We set the learning rate of the generator to 0.005, the hidden dim of the generator to 512, and the localize feature extractor to False. To diversify the output of the generator, the authors introduce a noise vector to the generator, we set the noise dim to 512.
    \item For Ditto, the client adjusts the preferences of the personalized model in the global model and the local model through the hyperparameter $\lambda$, we set the hyperparameter $\lambda$ to 0.001.
    \item For FedBABU, FedBABU fine-tunes the classifier of the global model for clients; we set the fine-tuning epochs to 10.
    \item For FedALA, we set the parameter for random select parameters rate to 0.8, the applying ALA on higher layers number to 1.
    \item For FedPake, we set the parameter for the threshold of coefficient of variation $\lambda$ to 0.2, and the threshold of similarity $\delta$ to 0.2, specially, for CIFAR-10/100, and Tiny-ImageNet, we set the number of Micro-Class $C$ to 4, 5, and 8, and the number of Macro-Class $S$ to 4, 5, and 8, respectively.
\end{itemize}

\subsection{Training Test Procedure}
There are differences in the training test processes of traditional FL and personal FL, which we explain separately.
\begin{itemize}
    \item \textbf{traditional FL}
    (1) \textbf{Server} initial global model and send it to selected clients; (2) \textbf{Client} initial local models by global model and test performance on private dataset; (4)\textbf{Client} training parameter on private dataset and send results to server; (4) \textbf{Server} collect local models and aggregate it to update global model, repeat step (1).
    \item \textbf{personal FL}
    (1) \textbf{Server} initial personalized models and send it to correspond clients; (2) \textbf{Client} initial local models by personalized models and test performance on private dataset; (4)\textbf{Client} training parameter on private dataset and send results to server; (4) \textbf{Server} collect local models and update personalized models, repeat step (1).
\end{itemize}
In standard form, the different of personal FL as followed: (1) Replace global model with personalized models designed specifically for each client; (2) Test performance by personalized models. Since our performance test criteria are a global model, we made the following changes for each personal FL baseline. Ditto and FedALA adopt the strategy of aggregating the local models to update the personalized models, therefore, we only need to change the testing objective to the aggregated result. For FedBABU, since the method does not aggregate local models, we directly test its personalized models. In conclusion, we test global models for all FL methods except FedBABU.

\section{FedPake Visualization}
\subsection{Hyperparameter Effect}
\begin{figure*}[ht]
    \centering
    \includegraphics[width=1\linewidth]{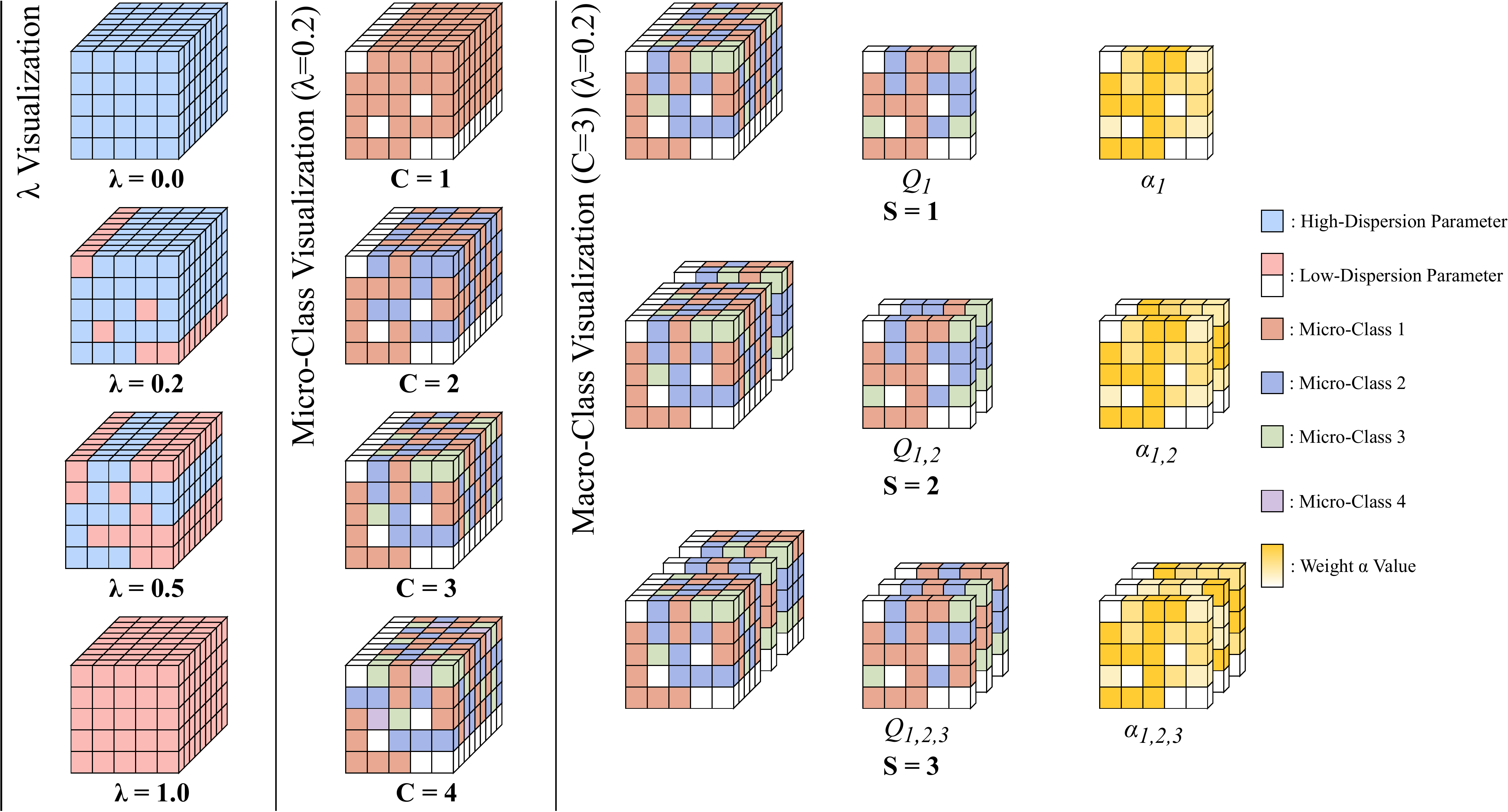}
    \caption{\textbf{The effect of Hyperparameters on FedPake,} the parameter is $5 \times 5$ convolution kernel (Length and height of a cuboid) and the number of client is 9 (Width of a cuboid). The hyperparameters of FedPake are affected by others, therefore, we indicate other hyperparameter values at the header of each section}
    \label{fig:FedPake_in_Param}
\end{figure*}

\begin{figure*}[ht]
    \centering
    \includegraphics[width=1\linewidth]{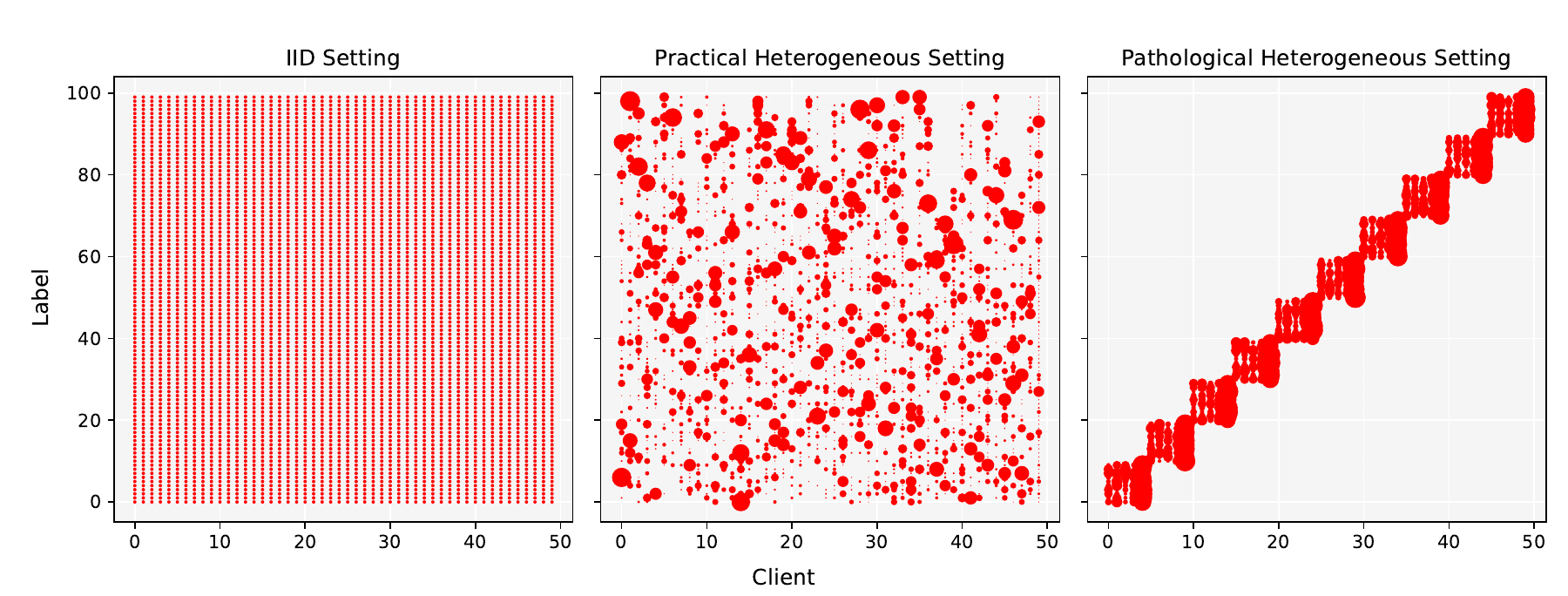}
    \caption{\textbf{The distribution of CIFAR-100 data under the IID setting, pathological setting, and practical setting}, in which the number of clients is 50. The size of the circle represents the number of samples.}
    \label{fig:label distribution(1)}
\end{figure*}

\begin{figure*}[ht]
    \centering
    \includegraphics[width=1\linewidth]{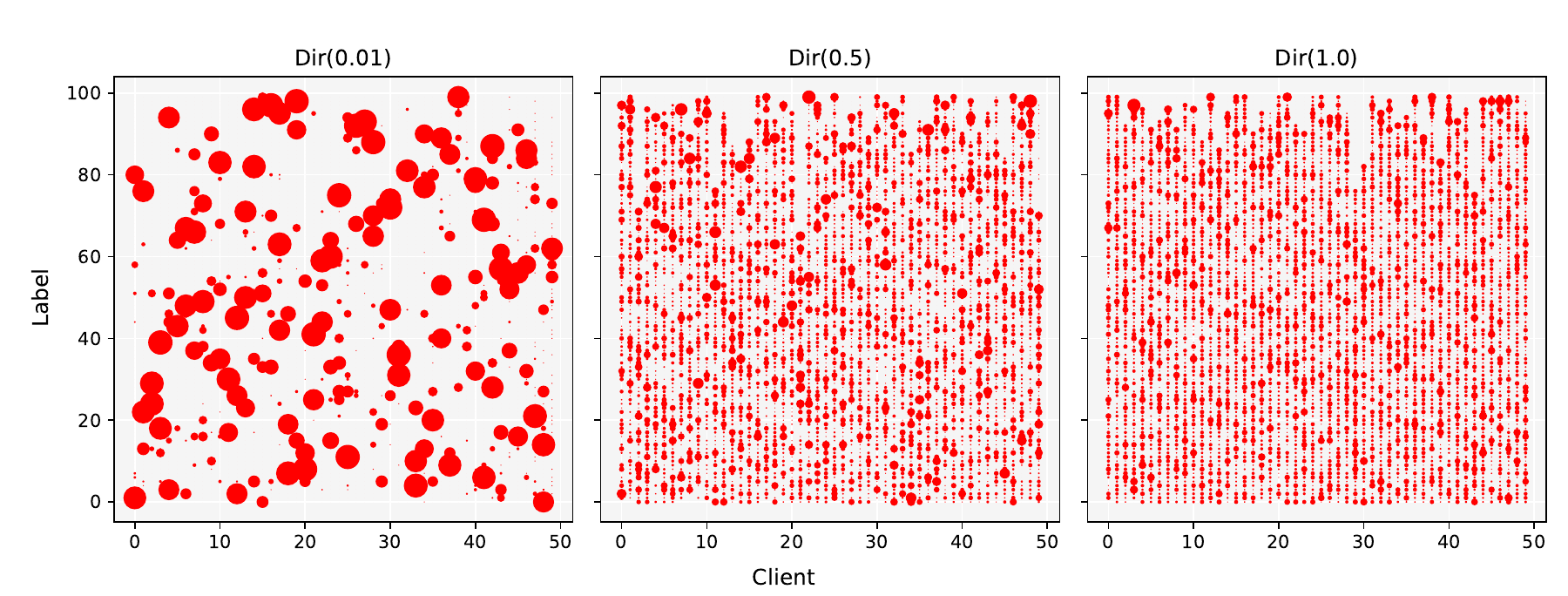}
    \caption{\textbf{The distribution of CIFAR-100 data under the $\operatorname{Dir(0.01)}$, $\operatorname{Dir(0.5)}$, $\operatorname{Dir(1.0)}$ heterogeneous setting}, which the number of clients is 50. The size of the circle represents the number of samples.}
    \label{fig:label distribution(2)}
\end{figure*}

Hyperparameter settings determine the performance of FedPake, so we present a visualization of FedPake under different hyperparameter settings. As shown in Figure \ref{fig:FedPake_in_Param}, $\, \lambda$ affects the parameter division stage; when $\lambda = 1.0$, FedPake does not function and appears similar to FedAVG. $C$ and $S$ affect the Micro-Class stage and Macro-Class stage, respectively. A higher $C$ means that the parameters will be divided into diverse categories, enhancing the differences between clients and providing more clustering possibilities, while $S$ limits the number of clusters.

\end{document}